\documentclass{article}
 \PassOptionsToPackage{numbers, compress}{natbib}

 \usepackage[preprint]{tackling_climate_workshop_style}

\usepackage[utf8]{inputenc} 
\usepackage[T1]{fontenc}    
\usepackage{hyperref}       
\usepackage{url}            
\usepackage{booktabs}       
\usepackage{amsfonts}       
\usepackage{nicefrac}       
\usepackage{microtype}      
\usepackage{graphicx}
\usepackage{subcaption}

\title{Harnessing Diverse Data for Global Disaster Prediction: A Multimodal Framework}

\author{%
  Gengyin Liu \\
  Department of Computer Science\\
  University of the Chinese Academy of Science\\
  Beijing, China \\
  \texttt{liugengyin20@mails.ucas.ac.cn} \\
  \And
  Huaiyang Zhong \\
  Department of Industrial and Systems Engineering\\
  Virginia Tech\\
  Blacksburg, VA , 24060\\
  \texttt{hzhong@vt.edu} \\
}

\begin{document}

\maketitle

\begin{abstract}
As climate change intensifies, the urgency for accurate global-scale disaster predictions grows. This research presents a novel multimodal disaster prediction framework, combining weather statistics, satellite imagery, and textual insights. We particularly focus on "flood" and "landslide" predictions, given their ties to meteorological and topographical factors. The model is meticulously crafted based on the available data and we also implement strategies to address class imbalance.  While our findings suggest that integrating multiple data sources can bolster model performance, the extent of enhancement differs based on the specific nature of each disaster and their unique underlying causes.  

\end{abstract}

\section{Introduction}
On the 10th of September 2023, Tropical Storm Daniel caused unprecedented flooding in Libya, resulting the loss of thousands of lives and many more reported missing. Natural disasters profoundly impacts communities, economies, and infrastructure, making the prediction vital for preparedness and mitigation \cite{allen2002constraints}. As climate changes and urban areas expand, the need for holistic, global-scale predictions becomes paramount. While single-modal methods can capture specific aspects, such as how landslides are influenced by weather, they might overlook other critical factors like soil type and geographical topography. Transitioning to a multimodal approach, which considers multiple facets of disaster prediction, presents its own challenges. The broader the scope, the harder it becomes to obtain comprehensive data for each modality. Given these complexities, there's a noticeable gap in research dedicated to global-scale, multimodal disaster prediction.

Different from single-modal data, multimodal data is introduced to have better potential, for each modality can influence the training of each other \cite{baltruvsaitis2018multimodal}. 
Integrating diverse data types can provide valuable potentials in many applications \cite{liu2021multimodal,nguyen2017deep,cadene2019murel,ngiam2011multimodal}.
Weather stands as a paramount external factor, wielding significant influence in the facilitation of specific natural calamities. In numerous instances, its indispensability and substantial impact remain unequivocal. Concurrently, geographical information harbors the potential to elucidate latent triggers, particularly in the context of disasters profoundly shaped by topographical factors, exemplified by phenomena like landslides.

This study introduces a multimodal disaster prediction framework, integrating diverse data sources including weather information, geographical information in the form of satellite imagery, and textual descriptions. 
By synthesizing these varied sources of information, our approach provides a unique solution for predicting different types of natural disasters. The results emphasize the critical role of combining multiple data streams to enhance the accuracy and comprehensiveness of disaster predictions.

\section{Data preparation}
\label{section:data prep}
We simulate predictions for specific 2021 disasters using historical data. We draw from the Geocoded Disasters (GDIS) dataset by Buhaug and Rosvold \cite{GDIS} and the EM-DAT disaster dataset \cite{EMDAT} to assemble our datasets. These sources offer detailed disaster data, including locations and names. Of the eight disaster types in GDIS, "flood" is the most frequent, making it a significant focus due to its prevalence. Concurrently, "landslide" is closely linked with meteorological events and satellite imagery. Given these insights, we've chosen to concentrate on "flood" and "landslide" for our study.

Our dataset, reflecting the temporal and spatial patterns of natural disasters, can exhibit an imbalance between positive and negative samples without proper filtering. As noted by Zeng and Bertimas (2023) \cite{zeng2023global}, the dataset focuses on specific urban areas. Only regions within a 100 km radius of cities with at least two recorded disasters (floods or landslides) from 1960 to 2018, as per GDIS, are included. This choice stems from the belief that such cities, due to their geography and climate, are prone to recurring disasters. Despite this approach, a significant imbalance persists, necessitating further adjustments during training.

Given this premise, we will collect data in three distinct categories for each selected city to construct our preliminary dataset, as detailed in the subsequent subsections.

\subsection{Statistical weather information}
To discern a sequential trend in weather variations, we gathered weather data spanning the past five years for each designated city. Given our objective to forecast disasters in 2021, the weather data pertains to the annual intervals from 2016 to 2020. Utilizing location data from the GDIS dataset, we initiated queries to \emph{VisualCrossing}, a repository for historical weather statistics \cite{Visual_Crossing_Corporation_undated-le} . We extracted the most representative features: \textit{(average) Temperature, Dew Point, Relative Humidity, Wind Speed \emph{and} Precipitation}. 

\subsection{Text description}
For a comprehensive understanding of the geographical, climatic, and hydrological attributes of each city in our dataset, we obtained detailed textual descriptions from OpenAI services, specifically querying the "text-davinci-002" model in the GPT-3 series via the OpenAI API \cite{davinci-002}. The prompts are listed below, where \emph{name} is the name of the city and the respective country, extracted from GDIS: [1] Describe the terrain of \emph{name}, [2] Describe the climate of \emph{name}, [3] What's the altitude of \emph{name}?, and [4] Describe the hydrology features of \emph{name} specifically.

\subsection{Satellite imagery}
 The likelihood of occurrences such as floods or landslides is intrinsically influenced by the geographical layout, terrain, and vegetation of a region. To gain a more nuanced understanding of the cities under study, we sourced supplementary satellite imagery. Google Earth Engine provides valid Sentinel-2/SR remote sensing images for most locations since 2018.  However, prior to this period, a subset of our selected locations encountered image unavailability. In alignment with our 2021 disaster prediction objective, we obtained the satellite images with the least cloudy percentage for 2020. More specifically, for each image we restrict the area as a square with a side length of 70,000 meters centered on the city, and the resolution ratio is set to 150 meters to limit the file size. We then overlay bands \textit{B4, B8A, B12} to accentuate the vegetation and waterway distribution.

\section{Methodology}
\label{section:methodology}

Our raw data remains intact, and we employ various strategies to extract features and address class imbalance. The initial train/test split ratio for both disaster datasets is set at 0.3.

\subsection{Data processing}
\label{subsection:data proc}
\subsubsection{Weather statistics}
We transform each location's tabular weather statistics into a 25-dimensional vector. While LSTM is effective for longer sequences, as highlighted by \cite{10.1162/neco.1997.9.8.1735}, it may not be optimal for short-term and sparse weather data. In weather prediction tasks, LSTMs are typically used when there's a need to capture long-term dependencies, as demonstrated by Karevan et al. (2020) \cite{KAREVAN20201}.
\subsubsection{Textual features}
Considering the best performed architecture In line with the architecture proposed by Zeng and Bertimas in 2023 \cite{zeng2023global}, we utilize a submodel for extracting textual features. This sub-model combines the original \emph{DistilBert} model with a custom linear layer containing 32 neurons. We conduct a 3-epoch training process on this sub-model, using a randomly selected 100-sample subset from our dataset to enhance its sensitivity to specific disasters. During training, we freeze the parameters of DistilBert.
\subsubsection{Imagery representation}
We use a pre-trained classifier for remote sensing images to extract hydrological and climatological insights. Since standard classifiers, like those trained on ImageNet, aren't tailored for our satellite data, we adopt classifiers fine-tuned for remote sensing from an open-source project \cite{lsh1994}. Specifically, the ResNet34 model excels, so we integrate it with a 46-neuron linear layer to form our image processing sub-model.  construct our image processing sub-model. We train this sub-model using a random 100-sample image subset to establish associations between images and disaster outcomes. 

\subsubsection{Oversampling}
As previously mentioned, our datasets suffer from significant class imbalance. To address this issue, we explore various strategies and find that \emph{oversampling} is the most effective. After splitting the flood dataset, we employ the Synthetic Minority Over-sampling Technique (SMOTE) \cite{SMOTE} on the training set to generate 520 additional positive samples, which are then combined with the original 126 positive samples and 646 negative samples for training. In the case of landslides, we generate 447 additional positive samples, resulting in a training set comprising 459 positive and 459 negative samples.

\subsection{Complete model architecture}
\label{subsection:show arch}
\begin{figure}
  \centering
  \includegraphics[width=0.65\textwidth]{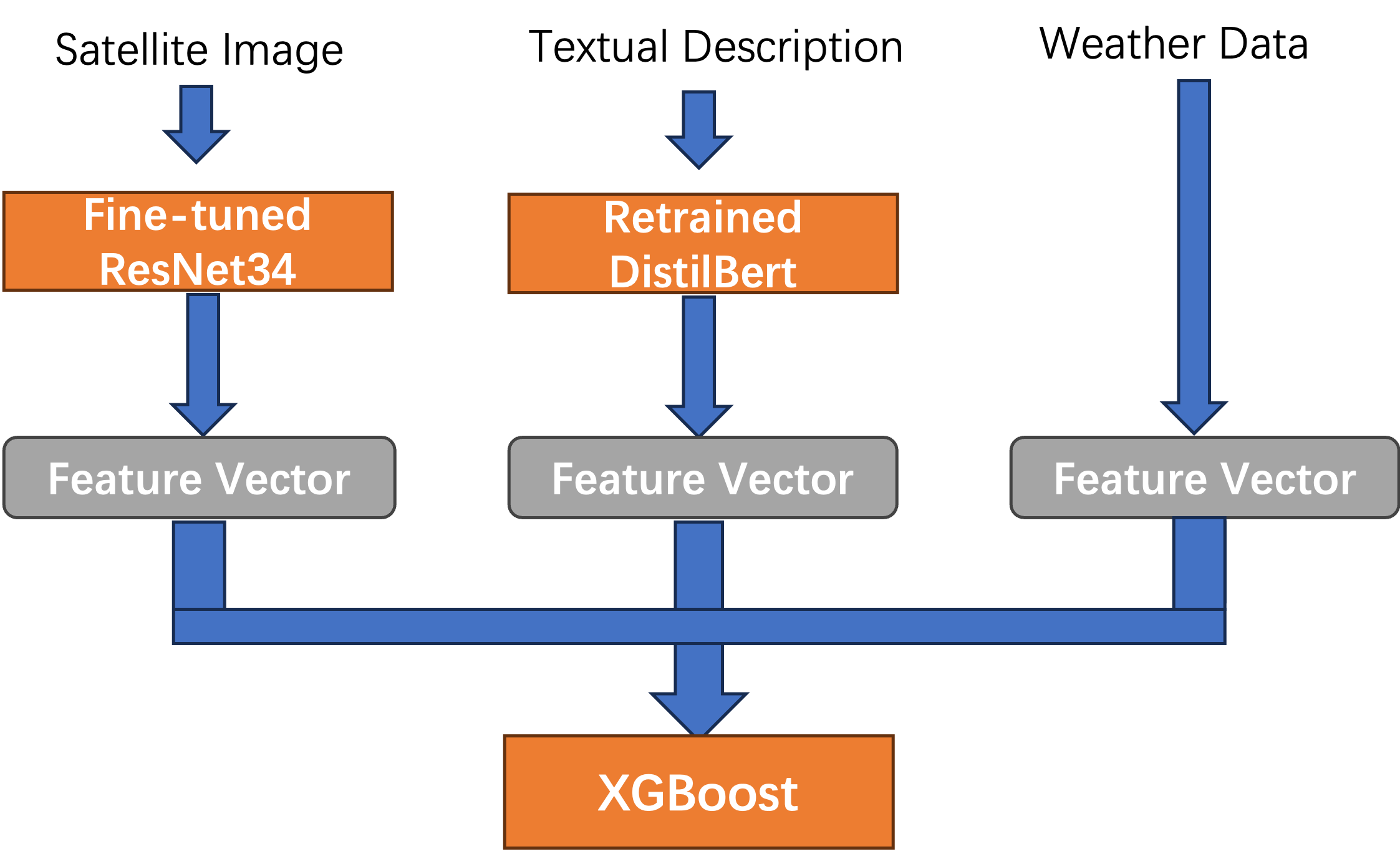}
  \caption{Complete process from raw data to prediction.}
  \label{fig:complete_model}
\end{figure}

As a multimodal prediction task, we collect data from various sources for each sample and process them through distinct pipelines. Subsequently, we aggregate all the features into a flat vector, which serves as the input for the XGBoost \cite{XGBoost} binary classifier. The XGBoost classifier undergoes 100 rounds of training with a learning rate of 0.1 and a maximum depth of 3 for both flood and landslide cases. An overview of our multimodal approach can be seen in Figure \ref{fig:complete_model}. 

\section{Results}
The performance metrics of various multimodal and single-modal architectures for predicting flood and landslide events are presented in Tables \ref{table:flood} and \ref{table:landslide}. Due to the class imbalance in our testing dataset, solely relying on accuracy can be misleading. Thus, we emphasize evaluation metrics like AUROC, F1 score, and balanced accuracy.

In general, models that incorporate statelite image consistently shows  improvements in both AUROC and F1 score across most scenarios. In the context of landslide prediction, the model encompassing weather, textual, and satellite images shows an improvement of 30\% in F1 score over the performance of the model relying solely on weather and textual data. Similarly, add statellite images into the text-onoy model also results in a 23.5\% increase in the AUROC score and a notable 46.8\% increase in the F1 score. It is important to note, however, that there are instances where the inclusion of satellite images does not yield a performance improvement. For example, in flood predictions, the model uses all three types of data has a slightly lower AUROC score than the one uses only weather and text data. This can be attributed to the significant influence of meteorological factors on floods, making image and text data more susceptible to noise. On the other hand, for disasters like landslides, which are more influenced by topographical factors and less by weather variations, the combination of image and text data proves to be more effective.

\begin{table}[h]
\footnotesize

\caption{Results for flood prediction.}
\resizebox{\linewidth}{!}{
\begin{tabular}{llllllll}
\hline
Metric   & Weather, Text, Image & Weather, Text & Weather, Image & Text, Image & Weather & Text    & Image     \\ \hline
AUROC      & 0.8176             & 0.8181        & 0.8129       & 0.6759    & 0.8512  & 0.5474  & 0.6710  \\
F1       & 0.4414             & 0.4277        & 0.4146       & 0.3297    & 0.4598  & 0.2246 
 & 0.3302  \\
Balanced accuracy & 71.31\%            & 71.22\%       & 70.33\%      & 62.00\%   & 75.75\% & 50.68\% & 63.02\% \\ \hline
\end{tabular}
}
\label{table:flood}
\end{table}

\begin{table}[h]
\footnotesize

\caption{Results for landslide prediction.}
\resizebox{\linewidth}{!}{
\begin{tabular}{llllllll}
\hline
Metric   & Weather, Text, Image & Weather, Text & Weather, Image & Text, Image & Weather & Text    & Image     \\ \hline
AUROC      & 0.9636             & 0.9556        & 0.9869       & 0.7889    & 0.9687  & 0.9384  & 0.6960  \\
F1       & 0.4000             & 0.3077        & 0.6667       & 0.3333    & 0.2857  & 0.3077  & 0.1905  \\
Balanced accuracy & 69.24\%            & 68.48\%       & 89.24\%      & 68.74\%   & 68.23\% & 68.48\% & 66.46\% \\ \hline
\end{tabular}
}
\label{table:landslide}
\end{table}

\section{Conclusion}
In conclusion, this research underscores the significance of diverse data in offering tailored insights for specific disaster types. Incorporating remote sensing imagery within the domain of multimodal disaster prediction framework enriches the analysis with vital geographical and hydrological details; while for disasters heavily influenced by meteorological factors, statistical weather data becomes indispensable. Our study advocates for a comprehensive strategy in global disaster prediction, seamlessly merging varied data types and harnessing the strengths of both computer vision and natural language processing techniques.  However, given the restricted temporal range and the suboptimal quality of available weather and imagery datasets, the application of more intricate models like LSTM, designed for time series data, becomes inherently challenging. This limitation curtails our ability to effectively monitor feature dynamics over time and pinpoint the primary triggers of diverse disasters across different geographical contexts.

\newpage

\bibliographystyle{unsrt}
\bibliography{references} 

\appendix

\end{document}